\pdfoutput=1

\documentclass[11pt]{article}

\usepackage[preprint]{acl}

\usepackage{times}
\usepackage{multirow}
\usepackage{latexsym}
\usepackage{makecell}
\usepackage{amsmath}
\usepackage{amssymb}
\usepackage{graphicx}
\usepackage{wasysym}
\usepackage{booktabs}
\usepackage{xcolor}
\usepackage{colortbl}
\usepackage{comment}
\usepackage[T1]{fontenc}

\usepackage[utf8]{inputenc}

\usepackage{microtype}

\usepackage{inconsolata}

\usepackage{graphicx}

\title{CORAL: Learning Consistent Representations across Multi-step Training with Lighter Speculative Drafter}

\author{Yepeng Weng, Dianwen Mei, Huishi Qiu\\ {\bf Xujie Chen, Li Liu, Jiang Tian, Zhongchao Shi} \\
AI Lab, Lenovo Research \\ 
\small{\texttt{\{wengyp1, meidw1, qiuhs1, chenxj23, liuli16, tianjiang1, shizc2\}@lenovo.com}}}

\begin{document}
\maketitle
\begin{abstract}

Speculative decoding is a powerful technique that accelerates Large Language Model (LLM) inference by leveraging a lightweight speculative draft model. However, existing designs suffers in performance due to misalignment between training and inference. Recent methods have tried to solve this issue by adopting a multi-step training strategy, but the complex inputs of different training steps make it harder for the draft model to converge. To address this, we propose CORAL, a novel framework that improves both accuracy and efficiency in speculative drafting. CORAL introduces Cross-Step Representation Alignment, a method that enhances consistency across multiple training steps, significantly improving speculative drafting performance. Additionally, we identify the LM head as a major bottleneck in the inference speed of the draft model. We introduce a weight-grouping mechanism that selectively activates a subset of LM head parameters during inference, substantially reducing the latency of the draft model. We evaluate CORAL on three LLM families and three benchmark datasets, achieving speedup ratios of 2.50$\times$-4.07$\times$, outperforming state-of-the-art methods such as EAGLE-2 and HASS. Our results demonstrate that CORAL effectively mitigates training-inference misalignment and delivers significant speedup for modern LLMs with large vocabularies.
\end{abstract}

\section{Introduction}

Large Language Models (LLMs), such as GPT \cite{DBLP:journals/corr/abs-2303-08774} and Llama series~\cite{touvron2023llamaopenefficientfoundation, touvron2023llama2openfoundation, grattafiori2024llama3herdmodels}, have demonstrated exceptional capabilities in various natural language processing tasks. However, achieving stronger model performance often depends on increasing the number of model parameters~\cite{kaplan2020scalinglawsneurallanguage, hoffmann2022trainingcomputeoptimallargelanguage}, which leads to higher costs in both training and inference. Thus, achieving strong performance while maintaining quick response is a crucial part in LLM implementations. Under common hardware conditions, transformer decoder-based LLMs are memory-bound \cite{DBLP:conf/nips/DaoFERR22}, which means that the generation speed is mainly determined by memory access and bandwidth, rather than arithmetic computations. This allows for the acceleration of generation using speculative decoding \cite{chen2023acceleratinglargelanguagemodel, DBLP:conf/icml/LeviathanKM23}. The general idea of speculative decoding is to utilize one or multiple lightweight draft models to predict the output of target LLM for several upcoming timesteps, and then verify the drafted predictions in parallel using the target model. The memory-bound characteristic guarantees that the parallel verification of multiple tokens does not incur a significant increase in latency compared to generating a single token.

\begin{figure}[tbp]
\centering
\includegraphics[width=7.5cm]{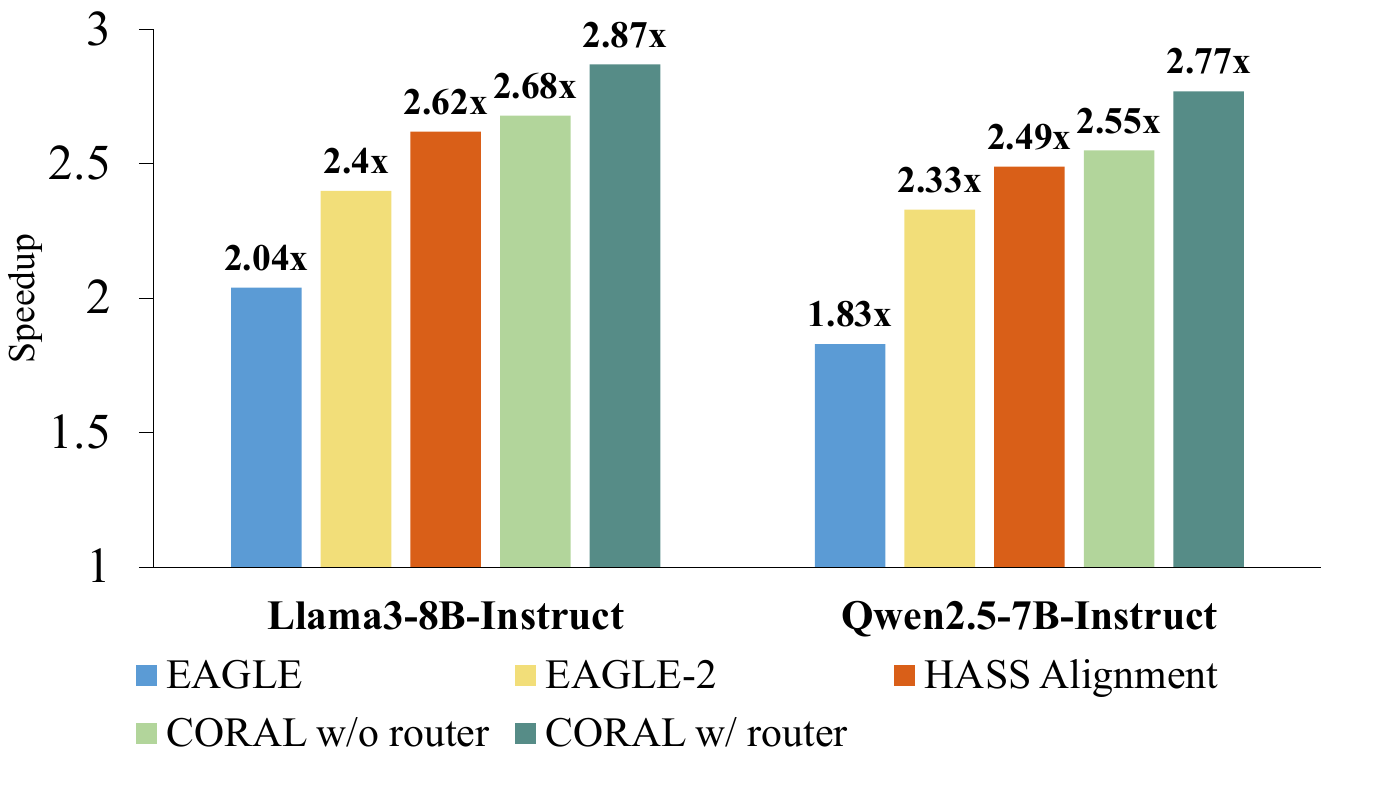}
\caption{Speedup ratios of different methods on Llama3-8B and Qwen2.5-7B at temperature=0, averaging on MT-bench, HumanEval, and GSM8K datasets. We present full results in Table \ref{tab:full_result} and this chart is only a subset of all comparisons.}
\label{fig:start}
\end{figure}

\begin{table*}[t]
    \centering
    \begin{tabular}{cccccccc}
    \toprule[0.7pt]
        \textbf{Model} & \textbf{Hidden} & \textbf{Inter. size} & \textbf{Vocab} & \boldmath{$W_{d}$}\ /\ $W_{t}$ & \boldmath{$L_{d}$}\ /\ $L_{t}$  \\ \midrule
        Llama2-7B & 4096 & 11008 & 32000 & 350M/6301M(5.6\%) & 1.36ms/23.65ms(5.8\%)\\ 
        Llama3-8B & 4096 & 14336 & 128256 & 741M/7157M(10.4\%) & 2.58ms/26.06ms(9.9\%)\\ 
        Qwen2.5-7B & 3584 & 18944 & 152064 & 767M/6743M(11.4\%) &  2.69ms/24.58ms(10.9\%)\\ \bottomrule[0.7pt]
    \end{tabular}
    \caption{Parameters and latencies of Llama3-8B, Llama2-7B, and Qwen2.5-7B draft and target models. $W_{d}$, $W_{t}$ and $L_{d}$, $L_{t}$ denote the parameter counts and latency of draft and target model. In the table, M represents 1024$\times$1024. Parameters of the embedding layer are not calculated because they do not participate in general matrix multiplication (GEMM). Latencies are tested with one token on a single NVIDIA A6000 GPU.}
    \label{tab:ldlt}
\end{table*}

Recently, autoregressive draft models, such as EAGLE \cite{li2024eagle}, have received widespread attention for their excellent speedup performance. For training, EAGLE uses not only the output tokens but also the last hidden states from target LLM as input to the draft model, while during the drafting phase, the draft model uses its own hidden states from the previous timestep, which may contain biases. This misalignment leads to a decrease in the prediction accuracy of the draft model. HASS \cite{zhang2024learningharmonizedrepresentationsspeculative} proposes a multi-step training strategy, where the hidden states output by the draft model are fed back into itself multiple times during training, allowing the draft model to learn the feature distribution of the inference phase. In Section \ref{section:preliminaries} we will provide more detailed discussions on them.

\begin{figure}[t]
    \centering
    \includegraphics[width=7.5cm]{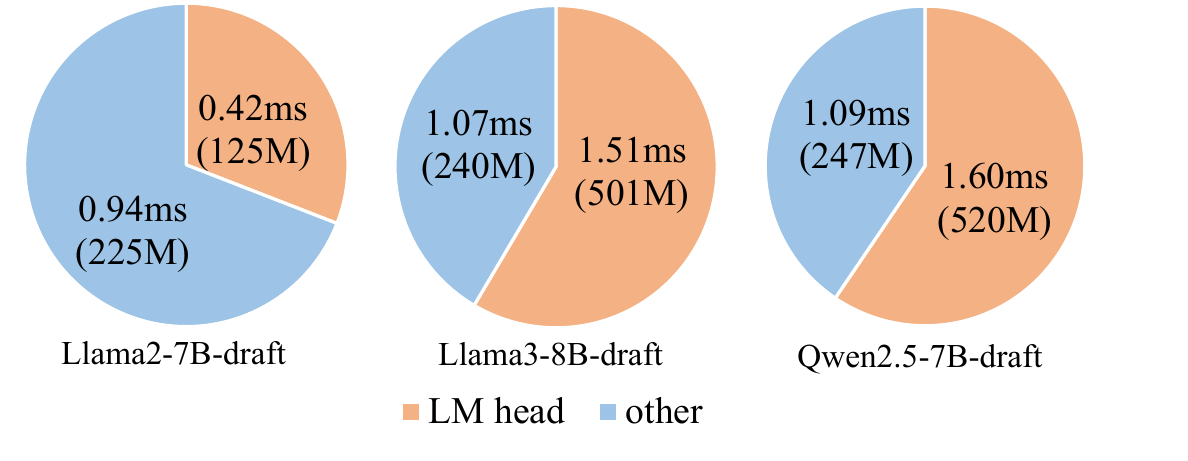}
    \caption{Parameters and latencies of Llama3-8B, Llama2-7B, Qwen2.5-7B draft model. For a model with large vocabulary, the LM head takes the majority of the drafting latency.}
    \label{fig:pie}
\end{figure}

Although HASS exhibits impressive performance, there are still some limitations to multi-step training. Specifically, their design causes the input features at differrent training steps to vary, which might be challenging for a lightweight draft model to adapt to. The discrepancy of each training step may also introduce potential gradient conflicts. Furthermore, modern LLMs are increasingly moving towards large vocabularies to obtain better performance \cite{tao2024scalinglawsvocabularylarger}. For example, previous model such as Llama2 has a small vocabulary size of only 32000 \cite{touvron2023llama2openfoundation}, while the vocabulary size of Llama3 \cite{grattafiori2024llama3herdmodels} is 128256, and that of Qwen2.5 \cite{qwen2.5} is 152064. Such large vocabularies lead to an increase in the parameter size of the Language Model head (LM head), resulting in increased overhead of drafting, which is presented in Table \ref{tab:ldlt}. As demonstrated in Figure \ref{fig:pie}, the heavy LM head could potentially dominate the latency of draft model. However, few studies have focused on this aspect.

In this paper, we introduce CORAL (learning COnsistent Representations Across multi-step training with Lighter speculative drafter), a speculative decoding method that improves the alignment between the draft model and the target model while maintaining high drafting speed. We first propose Cross-Step Representation Alignment (CSRA), which leverages the idea of contrastive learning to enforce consistency among the output features of each training step. The constraint on features makes them more stable, and thus improves the training efficiency and the performance of the draft model. Furthermore, by grouping the LM heads, we significantly reduce the activated parameters of the draft model with large vocabulary size, thereby decreasing the wall time of speculative decoding.

We evaluate acceleration capability of CORAL on multi-turn conversation, code generation, and mathematical reasoning tasks using the MT-Bench, HumanEval and GSM8K datasets, respectively. The results show that our method achieves 2.50$\times$-4.07$\times$ speedup over vanilla decoding at a temperature of 0, surpassing state-of-the-art methods such as EAGLE-2 and HASS.

\begin{figure*}[t]
    \centering
    \includegraphics[width=15.8cm]{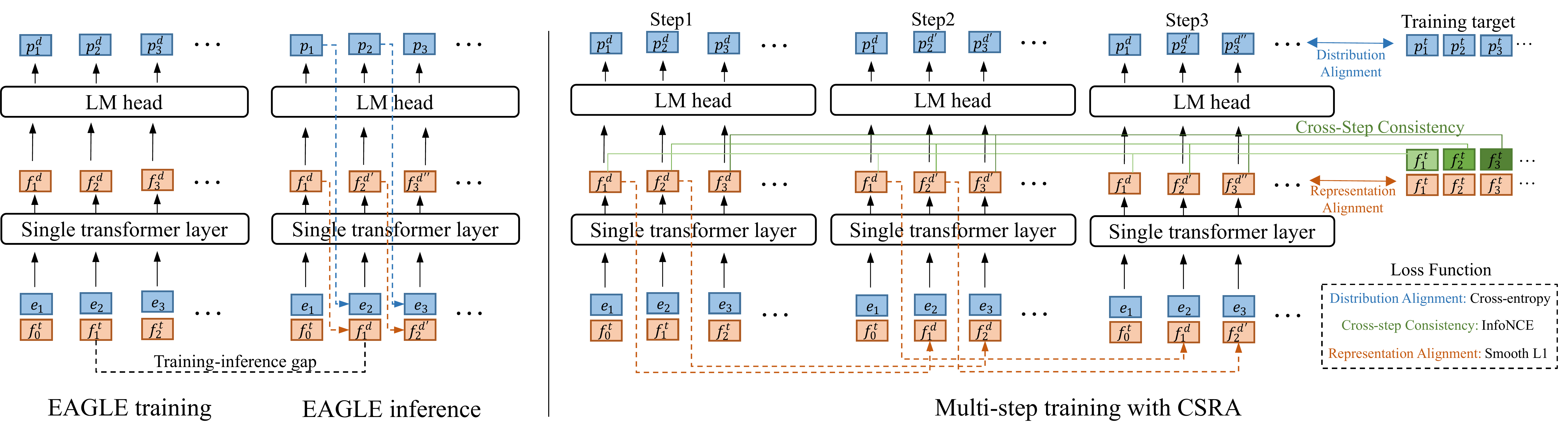}
    \caption{Demonstration of EAGLE training / inference and multi-step training with CSRA. $f$ denotes feature and $e$ denotes embedding. Superscripts indicate the source of the variable, with $t$ and $d$ denoting the target model and draft model. Subscripts index the position of a feature or embedding. For example, $f_{3}^{t}$ means the feature in position 3 and comes from the target model. For multi-step training, we use apostrophes to distinguish the outputs of different training steps. Specifically, we denote the output feature of step 1 as $f^{d}$, and for step 2 and 3 we use $f^{d'}$ and $f^{d''}$, respectively. Compared to HASS, CSRA introduces additional constraints on feature consistency. The training target is applied at each step, and we only illustrate it once for the sake of clarity.}
    \label{fig:pipeline} 
\end{figure*}

Our key contributions can be summarized as follows.
\begin{enumerate}
\item We propose Cross-Step Representation Alignment, a technique that enables the draft model to learn consistent representations across multiple timesteps.
\item We find that the vocabulary size can significantly influence the latency of the draft model, and propose a novel method which selectively activates a subset of LM head parameters during inference using a router.
\item CORAL achieves speedup ratios of 2.50$\times$-4.07$\times$ on various LLMs and datasets, outperforming existing speculative decoding methods such as EAGLE-2 and HASS.
\end{enumerate}

\section{Preliminaries}
\label{section:preliminaries}
In this section, we provide some background information related to speculative decoding and review some existing methods, including EAGLE and HASS.
\subsection{Speculative Decoding}
\label{section:sd}

Speculative decoding \cite{chen2023acceleratinglargelanguagemodel, DBLP:conf/icml/LeviathanKM23} aims to accelerate the generation speed of autoregressive LLMs. Vanilla speculative decoding employs a lightweight model (draft model) to generate a chain of candidate tokens for the next $\gamma$ timesteps, which are then verified in parallel by the original LLM (target model) and decide whether to accept them or not. Since the latency of LLM generation mainly lies in the memory access, parallel verification of multiple tokens does not significantly impact the latency of the target LLM, although the computational cost is multiplied.

The acceleration capability of speculative decoding is typically evaluated using two metrics: average acceptance length $\tau$ and the actual Speedup Ratio (SR). A drafting-verification cycle consists of one token provided by the target model and multiple candidates generated by the draft model over $\gamma$ time steps. The average acceptance length $\tau$ is defined as the number of new tokens generated in a single drafting-verification cycle.

Ideally, we can estimate the speedup ratio using $\tau$ and the latencies of draft and target model:
\begin{equation}
\label{eq:Speedup}
SR \approx \tau \times \frac {L_{t}'}{\gamma \times L_{d} + L_{t}},
\end{equation}
where $L_{t}$ and $L_{d}$ denote the latency of the target model and draft model, respectively. $L_{t}'$ denotes the latency for evaluating multiple tokens one time, it could be slightly different from $L_{t}$ depending on the hardware. Some additional overheads might also contribute to latency, such as comparing the probabilities of tokens from draft and target models to determine acceptance. However, since these overheads typically do not dominate the overall latency, it is a good choice to ignore them when estimating the speedup ratio.

From Equation \eqref{eq:Speedup} we can see the speedup ratio is primarily influenced by two factors: the alignment between the draft model and the target model, which mainly influences $\tau$, and the ratio of their latencies. Specifically, the lower the latency of the draft model and the better alignment between the two models, the higher the speedup ratio will be achieved by speculative decoding.

\subsection{EAGLE}

EAGLE \cite{li2024eagle} is a lightweight autoregressive draft model that leverages a single transformer layer identical to that of the target model. The LM head of draft model is reused directly from the target model, with its parameters frozen. EAGLE discovers that utilizing the feature (\emph{i.e.}, the last hidden states) of the target model can effectively enhance the alignment between the draft and target model. For training, the input of the draft model at position $s$ is the current token $t_{s}$ and the feature of the target model at position $s-1$. The token $t_{s}$ will first be transformed into embedding $e_{s}$, and then concatenated with the feature. A linear layer is adopted to reduce the dimensions before the single transformer layer.

The training target of EAGLE is to align the feature (regression) and probability distribution (classification) of the draft and target model. EAGLE uses smooth L1 as the regression loss and cross-entropy as the classification loss.

EAGLE selects multiple candidates at each timestep during drafting, resulting in a tree-shaped structure rather than a chain. Tree decoding offers more possible trajectories than chain decoding, leading to a higher acceptance length. EAGLE-2 \cite{li2024eagle2} improves the fixed tree structure to a dynamic one and achieves better performance.

\subsection{HASS}

HASS \cite{zhang2024learningharmonizedrepresentationsspeculative} addresses the inconsistency between the training and inference phases of EAGLE by introducing a multi-step training strategy. As demonstrated in Figure \ref{fig:pipeline}, EAGLE uses the feature of the target model for training, whereas in inference, the draft model uses its own feature. HASS solves this problem by feeding the output feature of draft model back into itself for multiple times. To expose the draft model to inference-time conditions during training, attention masks from different training steps require careful adjustment. HASS also incorporates other improvements on EAGLE, but they are orthogonal to multi-step alignment. In this paper, we focus mainly on HASS alignment, and all references to HASS in the remainder of this paper denote HASS alignment unless otherwise specified.

While HASS improves the accuracy of draft models in autoregressive generation, we argue that there are still unresolved issues due to the discrepancies between representations from multiple training steps (\emph{i.e.}, $f^{d}$, $f^{d'}$ and $f^{d''}$ in Figure \ref{fig:pipeline}). It is harder for the draft model to adapt to more complex inputs and the conflicting gradients from multiple steps may hinder convergence speed. 

\section{Method}
In this section, we first introduce Cross-Step Representation Alignment, a method designed to strengthen the alignment between the draft model and the target model. We then analyze the speedup ratio and identify the LM head of the draft model as a bottleneck. To address this issue, we propose the LM head router, a novel solution that aims to reduce the latency of the draft model.

\renewcommand{\dblfloatpagefraction}{.9}

\subsection{Cross-Step Representation Alignment}
Cross-Step Representation Alignment (CSRA) leverages the idea of contrastive learning \cite{DBLP:conf/cvpr/ChopraHL05, DBLP:conf/cvpr/SchroffKP15}. Specifically, in multi-step training, we treat the output features at the same position in a sentence as positive views of the same sample, while all other features are considered negative samples.

Assuming current training step is $t$, the output features of current step are $F_{t}\in\mathbb{R}^{B\times S\times D}$, where $B$, $S$, and $D$ represent the batch size, sequence length, and hidden dimension, respectively. Naturally, we regard them as $B \times S$ samples, and each sample has $t$ positive views, while all other features are considered negative samples.

\begin{figure}[t]
    \centering
    \includegraphics[width=7cm]{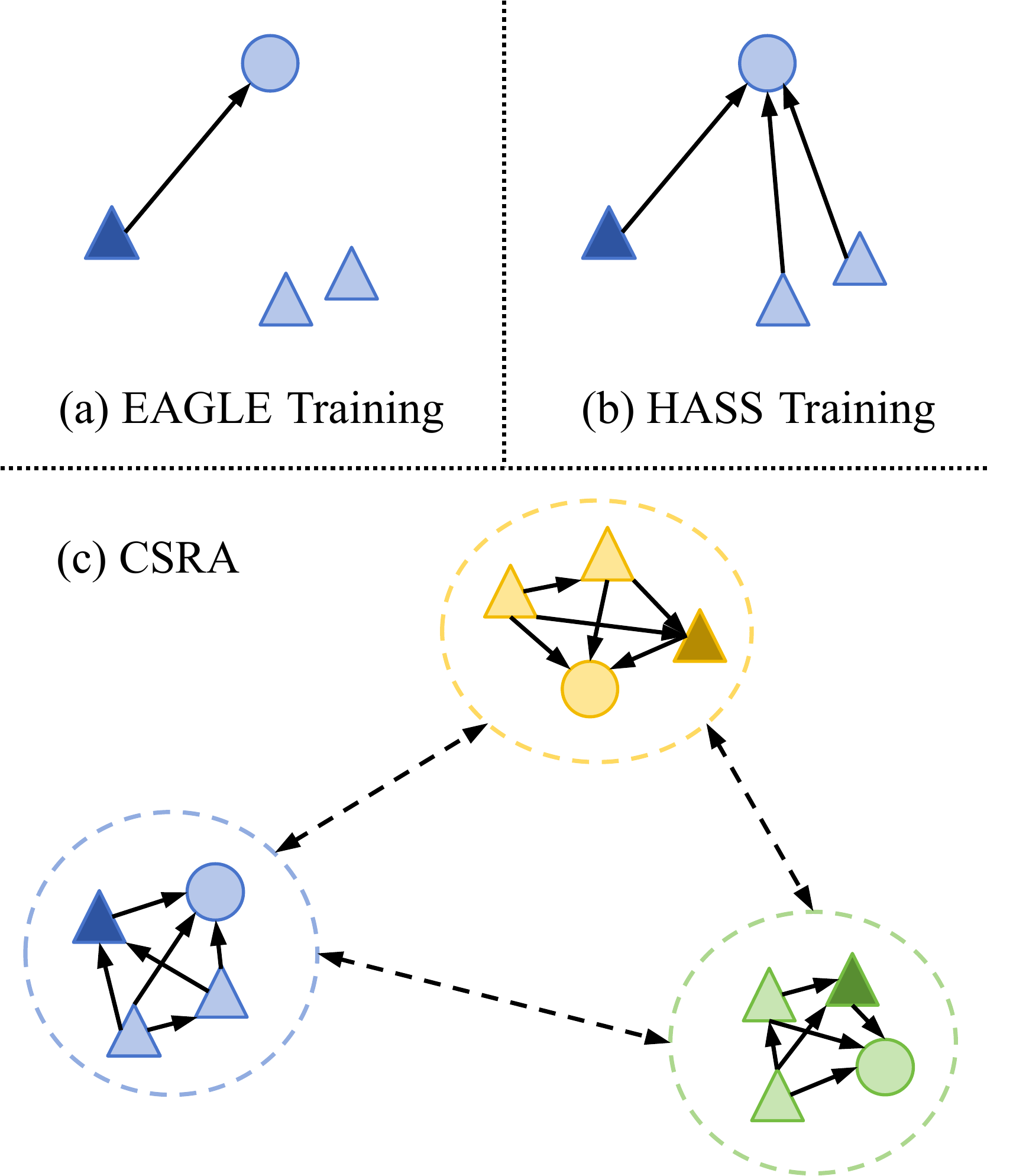}
    \caption{Comparison of EAGLE training, HASS training and CSRA. Here $\bigcirc$ denotes training target, $\bigtriangleup$ denotes output features from different steps. Triangles filled with darker colors represent the first step's output. Different colors represent outputs or targets of different positions. Optimization direction is marked as $\to$, and the dashed $\leftrightarrow$ means repulsion.}
    \label{fig:csra}
\end{figure}

For each output feature $f$ in current training step, our objective is to minimize its distance to other positive views while maximizing the distance to negative samples. To achieve this, we normalize the features and compute the InfoNCE loss \cite{DBLP:journals/corr/abs-1807-03748} as the objective function, which encourages the feature to be closer to its positive views and away from negative samples:

\begin{equation}
\label{eq:csra}
\mathcal{L}_{CSRA} = -\mathrm{log}\frac{\mathrm{exp}(\mathrm{sim}(q,f^{+})/\tau)}{\sum_{f\in F}\mathrm{exp}(\mathrm{sim}(q,f)/\tau)},
\end{equation}
where $q$ and $f^{+}$ denotes the query feature and positive views, and $F$ is the set of all features along with the targets. The similarity function $\mathrm{sim}(\cdot, \cdot)$ is defined as cosine similarity. Here $\tau$ is the temperature hyperparameter. Figure \ref{fig:csra} shows the differences between EAGLE / HASS training and CSRA.

The training loss can be defined as:
\begin{equation}
\label{eq:training_loss}
\mathcal{L} = w_{reg}\mathcal{L}_{reg}+w_{cls}\mathcal{L}_{cls}+w_{CSRA}\mathcal{L}_{CSRA},
\end{equation}
where $\mathcal{L}_{reg}$ and $\mathcal{L}_{cls}$ represent the regression loss and classification loss, respectively. Since $\mathcal{L}_{CSRA}$ primarily affects representation learning, we maintain $w_{cls}$ consistent with EAGLE and adjust another two weights according to different target models. For detailed parameter settings, please refer to Appendix \ref{sec:appendixA}.

\subsection{Estimation of Speedup Ratio}
\label{section:speedup}
As discussed in Section \ref{section:sd}, the generation speed is primarily constrained by memory bandwidth. 
Therefore, the theoretical latency $L_{theo.}$ in generation phase is proportional to the LLM's parameter count $W_{LLM}$:

\begin{equation}
\label{eq:ideal_latency}
L_{theo.} \propto W_{LLM}.
\end{equation}

However, this estimation is not always accurate due to the following factors: 1) Not all operators and computing graphs are fully optimized. 2) The latency of some element-wise operators (\emph{e.g.}, activation, norm) is not reflected in the parameter count. This issue is particularly noticeable for PyTorch, because it is not a framework optimized for inference.

Luckily, the draft model and target one share the same transformer structure, and the extra latency caused by the aforementioned factors is relatively consistent in both models. This allows us to estimate the wall time and speedup ratio of speculative decoding based on the parameters of draft model and target model:
\begin{equation}
\label{eq:ldlt}
\frac{L_{d}}{L_{t}} \approx \frac{W_{d}}{W_{t}},
\end{equation}

\begin{equation}
\label{eq:Speedup_param}
SR \approx \tau \times \frac {W_{t}}{\gamma \times W_{d} + W_{t}},
\end{equation}
where $W_{d}$, $W_{t}$ and $L_{d}$, $L_{t}$ denote the parameter counts and latency of draft and target model, respectively. Note that the embedding layer does not participate in general matrix multiplication (GEMM), therefore its parameters should not be included in latency estimation. Table \ref{tab:ldlt} presents the latencies and parameters of different LLMs, along with their corresponding draft models. The results suggest that estimating the latency ratio between the draft and target models based on their parameter counts is relatively accurate. Notably, for Llama3-8B and Qwen2.5-7B, the latency of draft model is approximately 10\% of that of target model. As the depth of drafting increases, the latency of draft model is expected to contribute significantly to the overall wall time.

Furthermore, it is also possible to estimate the latency of each component of the draft model based on their parameter count. As shown in Figure \ref{fig:pie}, in cases with large vocabularies, the latency of LM head accounts for a significant proportion of the total latency, which provides us with a valuable insight: If we can reduce the activated weights of the LM head, the overall speedup will be substantially improved.

\subsection{LM Head Router}
\label{section:router}
\begin{figure}[t]
\centering
\includegraphics[width=7.5cm]{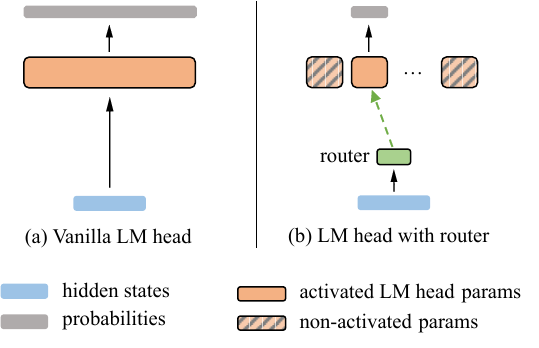}
\caption{Demonstration of LM head router in draft model. With the router, we only output probabilities of one or multiple subsets of vocabulary.}
\label{fig:router}
\end{figure}

As mentioned in Section \ref{section:speedup}, for draft models with large vocabularies, LM head constitutes the major part of drafting latency. We propose the LM head router, aiming to group the LM head and then activate only a subset of LM head parameters during drafting, as demonstrated in Figure \ref{fig:router}.

Assuming a LLM with a vocabulary size $V$, we divide the LM head equally into $N$ groups, each with a vocabulary size of $v=V/N$. We utilize a router to select which group to activate. The output of router can be outlined as follows:

\begin{equation}
\label{eq:father_head}
\begin{split}
p_{router}= \mathrm{Softmax}(W_{2}(\mathrm{act}(W_{1}h)+h)), \\
W_{2}\in\mathbb{R}^{N\times d},W_{1}\in\mathbb{R} ^ {d\times d},
\end{split}
\end{equation}
where $h$ denotes the hidden states of draft model, $d$ is the hidden size. 

Let $p(x)$, $q(x)$ denote the predicted and target distribution, and $p_\mathrm{{group}}(x^n)$ denote the probability distribution within a specific group $n$. After selecting a particular group, the softmax probability is calculated by logits in this group, independent of the logits in other groups.

Then the final distribution with router should be
\begin{equation}
\label{eq:final_distribution}
p(x)=p_\mathrm{{router}}(n)\cdot p_\mathrm{{group}}(x^{n}).
\end{equation}
For each group, $\sum p_\mathrm{{group}}(x^{n})=1$, and for router we have $\sum p_\mathrm{{router}}(n)=1$. Therefore, the final $p(x)$ is normalized.

The training target of LM head router is the sum of target probabilities in each group, namely $q_\mathrm{{router}}(n)=\sum q_\mathrm{{group}}(x^n)$. We use cross-entropy as the loss function:
\begin{equation}
\label{eq:loss_head}
\mathcal{L}_\mathrm{{router}}=-\sum{q_\mathrm{{router}}(n)\log p_\mathrm{{router}}(n)}.
\end{equation}

It is evident that, although the LM head router reduces the latency of the draft model, it comes at the cost of a slight decrease in acceptance length $\tau$ due to imperfect routing accuracy. Based on Equations \eqref{eq:ldlt} and \eqref{eq:Speedup_param}, the LM head router gets its best performance when 1) the LM head accounts for a significant portion of the latency of draft model 2) the latency ratio between the draft model and the target model is substantial. Therefore, we only apply the LM head router to models with large vocabularies (Qwen2.5, Llama3) and relatively small sizes (7B, 14B).

We adopt a two-stage training strategy, where we first train the draft model following the standard training procedure (either single-step or multi-step), and then fix the weights of draft model and train the router separately.  For further discussion, please refer to Appendix \ref{sec:appendixB}.

\section{Experiments}
In this section, we first introduce the experimental setup, then discuss the overall effectiveness of our method, and finally present the ablation studies on CSRA and LM head router.

\subsection{Experimental Setup}
\textbf{Target LLMs.} We choose Llama3-Instruct-8B/70B\cite{grattafiori2024llama3herdmodels}, Llama2-chat-7B/13B\cite{touvron2023llama2openfoundation} and Qwen2.5-Instruct-7B/14B\cite{qwen2.5} as our target models.

\noindent\textbf{Tasks.} We choose multiple datasets covering three tasks, including MT-Bench\cite{DBLP:conf/nips/ZhengC00WZL0LXZ23} for multi-turn dialogue, GSM8K\cite{DBLP:journals/corr/abs-2110-14168} for mathematical reasoning, and HumanEval\cite{DBLP:journals/corr/abs-2107-03374} for code generation. For 7B/14B models, experiments are conducted with batch size of 1 on a single NVIDIA A6000 48G GPU. For Llama3-70B-Instruct, we use 4$\times$A6000 GPUs due to memory requirements.

\noindent\textbf{Metrics.} Since CORAL is a lossless speculative decoding strategy, it is not necessary to measure the generation quality. For acceleration, we use two metrics to evaluate the performance: 
\begin{itemize}
    \item \textbf{Speedup Ratio}: the actual speedup ratio compared to vanilla decoding.
    \item \textbf{Acceptance Length \boldmath{$\tau$}}: the average number of new tokens generated per drafting-verification cycle.
\end{itemize}

\noindent\textbf{Comparisons.} We use vanilla decoding as the baseline (1.00$\times$) to measure the speedup ratio. We primarily compare CORAL with the latest lossless speculative decoding methods, including EAGLE, EAGLE-2, and HASS. Since EAGLE is already one of the fastest speculative decoding methods, we choose EAGLE as the speculative decoding baseline and do not compare with other methods with lower speedup ratios.

\noindent\textbf{Implementation.} Our implementation is based on the open source repositories of HASS\footnote{https://github.com/HArmonizedSS/HASS} and EAGLE-2\footnote{https://github.com/SafeAILab/EAGLE}, and the settings are primarily identical to those of them. All models are trained with ShareGPT dataset for 20 epochs with batch size of 2 per GPU. For HASS and CORAL, the default step for training is set to 3. Our system prompt for Llama3 is slightly different from that of EAGLE, please refer to Appendix \ref{sec:appendixC} for detailed discussion. For inference, we employ a tree depth of 6 and select 60 candidate tokens for all models. 

\begin{table*}[t]
    \centering
    \small
    \setlength{\tabcolsep}{0.15cm}
    \renewcommand{\arraystretch}{1.2}
    \begin{tabular}{cc|cc|cc|cc|c} \hline
        ~ & ~ & \multicolumn{2}{c|}{MT-bench} & \multicolumn{2}{c|}{HumanEval} & \multicolumn{2}{c|}{GSM8K} & Average \\ 
        ~ & ~ & \multicolumn{2}{c|}{$\tau$ / SR} & \multicolumn{2}{c|}{$\tau$ / SR} & \multicolumn{2}{c|}{$\tau$ / SR} & $\tau$ / SR\\ \hline
        model & method & T=0 & T=1 & T=0 & T=1 & T=0 & T=1  & T=0 \\ \hline
        \multirow{4}*{\makecell{L2-13B}} & EAGLE & 3.93/3.04$\times$ & 3.64/2.65$\times$ & 4.51/3.47$\times$ & 4.24/3.13$\times$ & 4.01/3.10$\times$ & 3.84/2.83$\times$ & 4.15/3.20$\times$ \\
        ~ & EAGLE-2 & 4.80/3.16$\times$ & 4.68/3.06$\times$ & 5.59/3.75$\times$ & 5.41/3.60$\times$ & 4.98/3.38$\times$ & 4.84/3.25$\times$ & 5.12/3.43$\times$ \\ 
        ~ & HASS & 5.20/3.42$\times$ & 5.02/3.26$\times$ & 5.99/4.01$\times$ & 5.79/3.86$\times$ & 5.32/3.60$\times$ & 5.24/3.51$\times$ & 5.50/3.68$\times$  \\ 
        \rowcolor{gray!10} \cellcolor{white} ~ & CORAL & \textbf{5.25/3.45$\times$} & \textbf{5.10/3.32$\times$} & \textbf{6.06/4.07$\times$} & \textbf{5.90/3.93$\times$} & \textbf{5.39/3.65$\times$} & \textbf{5.25/3.51$\times$} & \textbf{5.57/3.72$\times$}  \\ \hline
        \multirow{4}*{\makecell{L2-7B}} & EAGLE & 3.80/2.67$\times$ & 3.62/2.37$\times$ & 4.29/3.04$\times$ & 3.96/2.60$\times$ & 3.84/2.73$\times$ & 3.74/2.48$\times$ & 3.87/2.81$\times$ \\ 
        ~ & EAGLE-2 & 4.68/2.89$\times$ & 4.45/2.70$\times$ & 5.34/3.35$\times$ & 5.02/3.11$\times$ & 4.70/2.98$\times$ & 4.67/2.89$\times$ & 4.91/3.07$\times$  \\ 
        ~ & HASS & 5.02/3.09$\times$ & 4.77/2.88$\times$ & 5.71/3.58$\times$ & 5.35/3.30$\times$ & 5.11/3.25$\times$ & 4.99/3.10$\times$ & 5.28/3.31$\times$  \\ 
        \rowcolor{gray!10} \cellcolor{white} ~ & CORAL & \textbf{5.09/3.13$\times$} & \textbf{4.86/2.94$\times$} & \textbf{5.73/3.58$\times$} & \textbf{5.48/3.40$\times$} & \textbf{5.12/3.25$\times$} & \textbf{5.05/3.13$\times$} & \textbf{5.31/3.32$\times$}  \\ \hline
        \multirow{4}*{\makecell{L3-70B}} & EAGLE & 2.87/2.24$\times$ & 2.67/2.06$\times$ & 3.73/2.93$\times$ & 3.53/2.74$\times$ & 3.46/2.71$\times$ & 3.26/2.52$\times$ & 3.35/2.63$\times$ \\ 
        ~ & EAGLE-2 & 4.08/2.70$\times$ & 3.91/2.61$\times$ & 4.95/3.31$\times$ & 4.89/3.27$\times$ & 4.03/2.70$\times$  & 3.73/2.50$\times$ & 4.35/2.90$\times$  \\ 
        ~ & HASS & 4.10/2.71$\times$ & 4.00/2.65$\times$ & 5.23/3.49$\times$ & 5.10/3.40$\times$ & 4.12/2.76$\times$ & 3.83/2.56$\times$ & 4.48/2.99$\times$  \\ 
        \rowcolor{gray!10} \cellcolor{white} ~ & CORAL & \textbf{4.23/2.79$\times$} & \textbf{4.13/2.72$\times$} & \textbf{5.31/3.54$\times$} & \textbf{5.19/3.46$\times$} & \textbf{4.34/2.90$\times$} & \textbf{3.91/2.61$\times$} & \textbf{4.63/3.08$\times$}  \\ \hline
        \multirow{5}*{\makecell{L3-8B}} & EAGLE & 2.63/1.65$\times$ & 2.40/1.41$\times$ & 3.65/2.29$\times$ & 3.29/1.92$\times$ & 3.47/2.18$\times$ & 3.22/1.89$\times$ & 3.25/2.04$\times$ \\ 
        ~ & EAGLE-2 & 4.16/2.28$\times$ & 3.84/2.08$\times$ & 4.78/2.61$\times$ & 4.64/2.50$\times$ & 4.21/2.32$\times$ & 3.94/2.13$\times$ & 4.38/2.40$\times$  \\ 
        ~ & HASS & 4.48/2.45$\times$ & 4.12/2.21$\times$ & 5.31/2.89$\times$ & 5.12/2.76$\times$ & 4.56/2.51$\times$ & 4.18/2.28$\times$ & 4.78/2.62$\times$  \\ 
        \rowcolor{gray!10} \cellcolor{white} ~ & CORAL & \textbf{4.57/2.50$\times$} & \textbf{4.15/2.24$\times$} & \textbf{5.43/2.95$\times$} & \textbf{5.28/2.83$\times$} & \textbf{4.70/2.58$\times$} & \textbf{4.39/2.38$\times$} & \textbf{4.90/2.68$\times$}  \\ 
        \rowcolor{gray!20} \cellcolor{white} ~ & CORAL w/ r. & 4.26/\underline{\textbf{2.63$\times$}} & 3.92/\underline{\textbf{2.39$\times$}} & 5.22/\underline{\textbf{3.21$\times$}} & 5.03/\underline{\textbf{3.07$\times$}} & 4.42/\underline{\textbf{2.76$\times$}} & 4.12/\underline{\textbf{2.53$\times$}} & 4.63/\underline{\textbf{2.87$\times$}}  \\ \hline
        \multirow{5}*{\makecell{Q2.5-14B}} & EAGLE & 2.63/1.83$\times$ & 2.44/1.62$\times$ & 3.31/2.31$\times$ & 3.12/2.10$\times$ & 3.62/2.52$\times$ & 3.46/2.33$\times$ & 3.19/2.22$\times$ \\ 
        ~ & EAGLE-2 & 4.08/2.36$\times$ & 3.76/2.15$\times$ & 5.01/2.89$\times$  & 4.85/2.78$\times$ & 4.62/2.69$\times$ & 4.58/2.65$\times$ & 4.57/2.65$\times$  \\ 
        ~ & HASS & 4.52/2.59$\times$ & 4.12/2.35$\times$ & 5.50/3.18$\times$ & 5.37/3.07$\times$ & 5.03/2.92$\times$ & 4.91/2.83$\times$ & 5.02/2.90$\times$   \\ 
        \rowcolor{gray!10} \cellcolor{white} ~ & CORAL & \textbf{4.56/2.62$\times$} & \textbf{4.13/2.35$\times$} & \textbf{5.64/3.26$\times$} & \textbf{5.40/3.09$\times$} & \textbf{5.16/3.00$\times$}  & \textbf{5.12/2.95$\times$} & \textbf{5.12/2.96$\times$}  \\ 
        \rowcolor{gray!20} \cellcolor{white} ~ & CORAL w/ r. & 4.26/\underline{\textbf{2.74$\times$}} & 3.88/\underline{\textbf{2.46$\times$}} & 5.31/\underline{\textbf{3.44$\times$}} & 5.12/\underline{\textbf{3.28$\times$}} & 4.80/\underline{\textbf{3.14$\times$}} & 4.72/\underline{\textbf{3.05$\times$}} & 4.79/\underline{\textbf{3.11$\times$}} \\ \hline
        \multirow{5}*{\makecell{Q2.5-7B}} & EAGLE & 2.53/1.56$\times$ & 2.25/1.27$\times$ & 3.04/1.87$\times$ & 2.79/1.58$\times$ & 3.32/2.05$\times$ & 3.00/1.72$\times$ & 2.96/1.83$\times$ \\
        ~ & EAGLE-2 & 3.91/2.13$\times$ & 3.45/1.86$\times$ & 4.62/2.53$\times$ & 4.36/2.35$\times$ & 4.23/2.33$\times$ & 4.07/2.21$\times$ & 4.25/2.33$\times$  \\ 
        ~ & HASS & 4.15/2.26$\times$ & 3.65/1.96$\times$ & 4.96/2.71$\times$ & 4.74/2.55$\times$ & 4.53/2.49$\times$ & 4.35/2.35$\times$ & 4.55/2.49$\times$  \\ 
        \rowcolor{gray!10} \cellcolor{white} ~ & CORAL & \textbf{4.22/2.30$\times$} & \textbf{3.83/2.05$\times$} & \textbf{5.09/2.78$\times$} & \textbf{4.86/2.62$\times$} & \textbf{4.67/2.57$\times$} & \textbf{4.50/2.44$\times$} & \textbf{4.66/2.55$\times$}  \\ 
        \rowcolor{gray!20} \cellcolor{white} ~ & CORAL w/ r. & 4.02/\underline{\textbf{2.50$\times$}}  & 3.62/\underline{\textbf{2.21$\times$}} & 4.86/\underline{\textbf{3.05$\times$}} & 4.57/\underline{\textbf{2.81$\times$}} & 4.38/\underline{\textbf{2.76$\times$}} & 4.16/\underline{\textbf{2.58$\times$}} & 4.42/\underline{\textbf{2.77$\times$}}  \\ \hline
    \end{tabular}
    \caption{Acceptance lengths $\tau$ and speedup ratio (SR) of different methods on MT-bench, HumanEval, and GSM8K datasets with temperature $T \in \{0,1\}$. The best results are in \textbf{bold}, and some minor advantages may be obscured due to rounding. We also calculate the average $\tau$ and SR under $T=0$ for a more direct comparison. L2, L3, Q2.5 represents Llama2-Chat, Llama3-Instruct, and Qwen2.5-Instruct, respectively. As clarified in Section \ref{section:router}, we apply LM head router for relatively small LLMs with large vocabularies (denoted as CORAL w/ r.), such as Qwen2.5-7B/14B and Llama3-8B. For Llama2 series and Llama3-70B, we use CSRA only.}
    \label{tab:full_result}
\end{table*}

\subsection{Effectiveness and Ablation Studies}
\subsubsection{Effectiveness}
We present the acceptance lengths $\tau$ and speedup ratios of three datasets in Table \ref{tab:full_result}. The results show that CSRA achieves the best performance in both $\tau$ and speedup ratio (SR) in all experiments we have tested, surpassing EAGLE, EAGLE-2, and HASS. The advantages of CSRA are more pronounced for LLMs with larger vocabularies, whereas the benefits are less significant for earlier models such as Llama2. For LM head router, we set the group number to 16 and choose the top-2 groups for the best performance. Although the router sacrifices some acceptance length, the overall speedup ratio benefits from reduced latency and shows a considerable increase. 

\begin{table*}[t]
    \centering
    \small
    \setlength{\tabcolsep}{0.15cm}
    \renewcommand{\arraystretch}{1.2}
    \begin{tabular}{c|cc|cc|cc|cc}
    \hline
        ~ & \multicolumn{2}{c|}{MT-bench} & \multicolumn{2}{c|}{HumanEval} & \multicolumn{2}{c|}{GSM8K} & \multicolumn{2}{c}{Average} \\ \hline
        step & HASS & CSRA & HASS & CSRA & HASS & CSRA & HASS & CSRA \\ \hline
        2 & 4.41/2.41$\times$ & \textbf{4.53/2.48$\times$} & 5.24/2.86$\times$ & \textbf{5.35/2.90$\times$} & 4.50/2.47$\times$ & \textbf{4.60/2.52$\times$} & 4.72/2.58$\times$ & \textbf{4.83/2.63$\times$} \\
        3 & 4.48/2.45$\times$ & \textbf{4.57/2.50$\times$} & 5.31/2.89$\times$ & \textbf{5.43/2.95$\times$} & 4.56/2.51$\times$ & \textbf{4.70/2.58$\times$} & 4.78/2.62$\times$ & \textbf{4.90/2.68$\times$} \\
        4 & 4.46/2.44$\times$ & \textbf{4.58/2.51$\times$} & 5.39/2.93$\times$ & \textbf{5.55/3.00$\times$} & 4.58/2.54$\times$ & \textbf{4.70/2.57$\times$} & 4.81/2.64$\times$ & \textbf{4.94/2.69$\times$} \\
    \hline
    \end{tabular}
    \caption{Acceptance length and speedup ratio of Llama3-8B under different alignment steps.}
    \label{tab:CSRA&HASS}
\end{table*}

\subsubsection{Ablation Study on CSRA}
We conduct a more detailed comparative analysis with HASS under different training steps. According to HASS, further increases in the number of training steps (\emph{i.e.,} training steps $\geq$ 5) do not necessarily lead to improvements in acceptance length. Therefore, we focus our comparison on the cases where the number of training steps is set to 2, 3 (default), and 4.

\begin{figure}[h]
\centering
\setlength{\abovecaptionskip}{0.0cm}
\includegraphics[width=7.5cm]{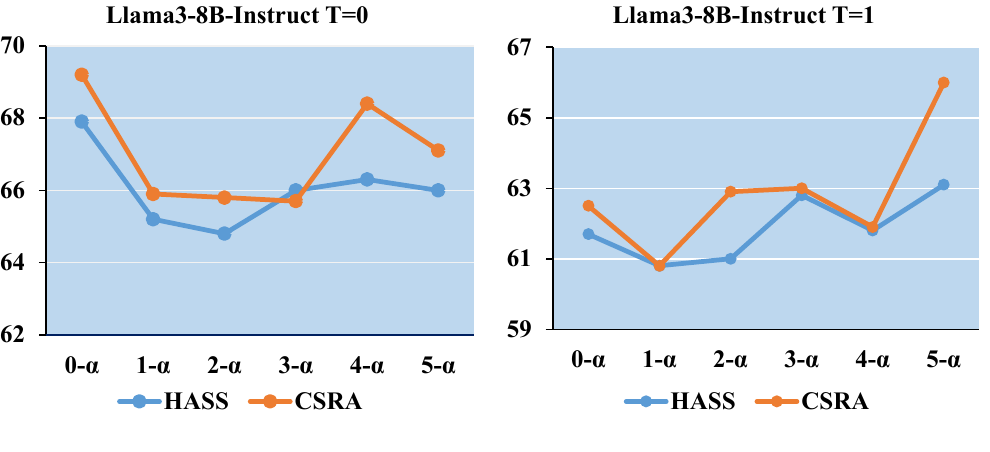}
\caption{Acceptance rates in MT-bench dataset. Here n-$\alpha$ denotes the acceptance rate of the n-th token.}
\label{fig:coral}
\end{figure}

The results summarized in Table \ref{tab:CSRA&HASS} demonstrate that CSRA consistently outperforms HASS with the same number of training steps. Furthermore, from the perspective of training cost, CSRA achieves a performance comparable to HASS (step=4) using only 2 training steps, demonstrating that CSRA offers a substantial advantage in terms of training efficiency. We also compare the acceptance rates $\alpha$ of HASS and CSRA at different timesteps during inference, as shown in Figure \ref{fig:coral}. The results show that CSRA generally outperforms HASS at different timesteps.

\subsubsection{Ablation Study on LM Head Router}
The LM head router has two hyperparameters: the total number of groups $N$, and the number of top-$n$ groups to activate during inference. A larger group number, although leading to activating fewer parameters, would increase the difficulty of training and damage accuracy. Similarly, how many groups to activate is also a trade-off between speed and accuracy. We perform a grid search over these two hyperparameters in the MT-bench dataset with Llama3-8B, and the results are shown in Table \ref{tab:speedup_with_router}.

\begin{table}[htb]
    \centering
    \setlength{\tabcolsep}{0.12cm}
    \scalebox{0.9}{
    \begin{tabular}{ccccccc}
    \toprule[0.7pt]
        \multicolumn{7}{c}{CORAL T=0} \\ \midrule
        N & top1 & top2 & top3 & top4 & top6 & top8 \\ \midrule
        N/A & 2.50$\times$ & - & - & - & - & - \\
        4 & 2.60$\times$ & 2.46$\times$ & - & - & - & - \\ 
        8 & 2.62$\times$ & 2.61$\times$ & 2.54$\times$ & - & - & - \\ 
        16 & 2.53$\times$ & \textbf{2.63$\times$} & 2.60$\times$ & 2.57$\times$ & - & - \\ 
        32 & 2.41$\times$ & 2.59$\times$ & 2.60$\times$ & 2.61$\times$ & 2.56$\times$ & - \\ 
        64 & 2.33$\times$ & 2.51$\times$ & 2.55$\times$ & 2.57$\times$ & 2.57$\times$ & 2.53$\times$ \\ \midrule
        \multicolumn{7}{c}{EAGLE-2 T=0} \\ \midrule
        N & top1 & top2 & top3 & top4 & top6 & top8 \\ \midrule
        N/A & 2.28$\times$ & - & - & - & - & - \\
        4 & \textbf{2.44$\times$} & 2.29$\times$ & - & - & - & - \\ 
        8 & 2.40$\times$ & 2.39$\times$ & 2.33$\times$ & - & - & - \\ 
        16 & 2.30$\times$ & 2.41$\times$ & 2.39$\times$ & 2.36$\times$ & - & - \\ 
        32 & 2.24$\times$ & 2.37$\times$ & 2.40$\times$ & 2.38$\times$ & 2.35$\times$ & - \\
        64 & 2.18$\times$ & 2.33$\times$ & 2.37$\times$ & 2.37$\times$ & 2.37$\times$ & 2.33$\times$ \\ \bottomrule[0.7pt]
    \end{tabular}
    }
    \caption{Speedup of Llama3-8B with LM head router on MT-bench dataset. We group the LM head parameters into $N$ groups and selectively activate top-$n$ of them. N/A denotes the results without LM head router.}
    \label{tab:speedup_with_router}
\end{table}

The results show that our method consistently yields significant improvements, regardless of whether multi-step training is employed. For CORAL, dividing the LM head into 16 groups and activating the top-2 groups during inference brings the best speedup performance.  Since the optimal setting may vary across different LLMs and cannot be easily estimated, we recommend empirical studies to identify the optimal configuration.

Let us discuss the effectiveness of the LM head router from another aspect using Llama3-8B. Since the number of activated LM head groups may vary due to tree decoding, we can estimate the latency of the draft model based on the ratio between acceptance length and speedup. Specifically, according to Table \ref{tab:ldlt}, the latency of the draft model is approximately 10\% that of the target model. Therefore, ideally (assuming $L_t' = L_t$ in Equation \ref{eq:Speedup}), the ratio between acceptance length and speedup should be 1.6, which means that within one drafting-verification cycle, the target model is invoked once while the draft model runs six times. However, in practical scenarios, the latency of the target model increases by approximately 19\% (from 26ms to 31ms) when inferring 60 tokens in parallel compared to generating a single token. Taking into account this factor, the actual ratio between acceptance length and speedup will increase to approximately 1.8.

Our experiments also confirm this estimation. Without the LM head router, the ratio between acceptance length and speedup is approximately $4.90/2.68 \approx 1.83$. In contrast, when the LM head router is adopted, this ratio decreases to $4.63/2.87 \approx 1.61$. This indicates that the average latency of the draft model is only $(0.6 - 0.22)/0.6 \approx 63\%$ of its original latency, demonstrating the efficacy of the LM head router.

\section{Related Work}
There has been a significant amount of work in accelerating LLMs. Some methods focus on reducing the number of parameters or memory access, such as low-bit quantization \cite{dettmers2022llmint88bitmatrixmultiplication, frantar2023gptqaccurateposttrainingquantization, DBLP:conf/icml/XiaoLSWDH23, DBLP:conf/mlsys/0002TTYCWXDG024}, and model distillation \cite{gu2024minillm, DBLP:conf/icml/KoKCY24, DBLP:conf/acl/Zhong00L0T24}. Recently, some studies have also explored activating only a subset of model parameters during inference to reduce memory access cost \cite{DBLP:conf/icml/DuHDTLXKZYFZFBZ22, DBLP:journals/jmlr/FedusZS22}. Speculative decoding \cite{chen2023acceleratinglargelanguagemodel, DBLP:conf/icml/LeviathanKM23} leverages the memory-bound nature of decoder-only LLMs and achieves lossless acceleration using a drafting-verification framework.

Research on speculative decoding has primarily focused on two areas: 1) drafter design, 2) verification strategy. For drafter design, Medusa \cite{DBLP:conf/icml/CaiLGPLCD24} attaches multiple heads to the original LLM and predict multiple subsequent tokens one time. Hydra \cite{ankner2024hydrasequentiallydependentdraftheads} improves Medusa by enhancing correlations between draft heads. Clover \cite{DBLP:journals/corr/abs-2405-00263} introduces an RNN-based draft head. Some methods utilize more information from target model to improve alignment, EAGLE \cite{li2024eagle} combines the output token and last hidden states of target LLMs to resolve the uncertainty in drafter's prediction. GLIDE \cite{DBLP:conf/icml/Du0XWY0LXNTY24} reuses the KV cache of target LLMs. For the verification strategy,  \citet{hu2024accelerated, sun2024blockverificationacceleratesspeculative} find that the acceptance length of speculative sampling is not optimal and take into account the probability of subsequent tokens. SpecInfer \cite{DBLP:conf/asplos/MiaoOZCWZWZYSSC24} proposes decoding tree for verification. Sequoia \cite{chen2024sequoiascalablerobusthardwareaware}, EAGLE-2 \cite{li2024eagle2}, and OPT-tree \cite{wang2024opttreespeculativedecodingadaptive} adopts a dynamic tree structure.

\section{Conclusion}
This paper proposes CORAL, an efficient speculative decoding method. We introduce Cross-Step Representation Alignment, which effectively mitigates training-inference misalignment and improves the accuracy of speculation. Additionally, we propose the LM head router, a plug-and-play module designed to reduce the latency of the draft model. We compare CORAL with other state-of-the-art methods on various LLMs and datasets, and the results show that CORAL surpasses existing methods, such as EAGLE-2 and HASS, demonstrating the effectiveness of our method.

\section*{Limitations}
There are mainly two limitations in this work.
Firstly, the introduction of CSRA loss may lead to a slight increase in regression loss, which results in a decrease in the acceptance length if the draft model is trained with single step. This issue can be addressed by multi-step training.
Secondly, adopting a large vocabulary is a trend in the development of modern LLMs, and our LM head router is specifically designed for LLMs with large vocabularies. It might not be suitable for models with small vocabularies, as the computational overhead of LM head is limited in the overall wall time of speculative decoding. In this case, the time saved by the draft model cannot compensate for the loss in acceptance length.

\section*{Acknowledgments}
We would like to thank Lenovo Model Factory team for providing computing resources. Special thanks to Xiaoyue Mi from the Institute of Computing Technology, Chinese Academy of Sciences, Penghui Yang from Nanyang Technological University, and Henry Zheng from Tsinghua University for their valuable suggestions during the writing of this paper.

\bibliography{custom}

\begin{thebibliography}{44}
\providecommand{\natexlab}[1]{#1}

\bibitem[{Ankner et~al.(2024)Ankner, Parthasarathy, Nrusimha, Rinard, Ragan-Kelley, and Brandon}]{ankner2024hydrasequentiallydependentdraftheads}
Zachary Ankner, Rishab Parthasarathy, Aniruddha Nrusimha, Christopher Rinard, Jonathan Ragan-Kelley, and William Brandon. 2024.
\newblock Hydra: Sequentially-dependent draft heads for medusa decoding.
\newblock \emph{arXiv preprint arXiv:2402.05109}.

\bibitem[{Cai et~al.(2024)Cai, Li, Geng, Peng, Lee, Chen, and Dao}]{DBLP:conf/icml/CaiLGPLCD24}
Tianle Cai, Yuhong Li, Zhengyang Geng, Hongwu Peng, Jason~D. Lee, Deming Chen, and Tri Dao. 2024.
\newblock Medusa: Simple {LLM} inference acceleration framework with multiple decoding heads.
\newblock In \emph{Proceedings of the International Conference on Machine Learning}.

\bibitem[{Chen et~al.(2023)Chen, Borgeaud, Irving, Lespiau, Sifre, and Jumper}]{chen2023acceleratinglargelanguagemodel}
Charlie Chen, Sebastian Borgeaud, Geoffrey Irving, Jean-Baptiste Lespiau, Laurent Sifre, and John Jumper. 2023.
\newblock Accelerating large language model decoding with speculative sampling.
\newblock \emph{arXiv preprint arXiv:2302.01318}.

\bibitem[{Chen et~al.(2021)Chen, Tworek, Jun, Yuan, de~Oliveira~Pinto et~al.}]{DBLP:journals/corr/abs-2107-03374}
Mark Chen, Jerry Tworek, Heewoo Jun, Qiming Yuan, Henrique~Pond{\'{e}} de~Oliveira~Pinto, et~al. 2021.
\newblock Evaluating large language models trained on code.
\newblock \emph{arXiv preprint arXiv:2107.03374}.

\bibitem[{Chen et~al.(2024)Chen, May, Svirschevski, Huang, Ryabinin, Jia, and Chen}]{chen2024sequoiascalablerobusthardwareaware}
Zhuoming Chen, Avner May, Ruslan Svirschevski, Yuhsun Huang, Max Ryabinin, Zhihao Jia, and Beidi Chen. 2024.
\newblock Sequoia: Scalable, robust, and hardware-aware speculative decoding.
\newblock \emph{arXiv preprint arXiv:2402.12374}.

\bibitem[{Chopra et~al.(2005)Chopra, Hadsell, and LeCun}]{DBLP:conf/cvpr/ChopraHL05}
Sumit Chopra, Raia Hadsell, and Yann LeCun. 2005.
\newblock Learning a similarity metric discriminatively, with application to face verification.
\newblock In \emph{Proceedings of the Conference on Computer Vision and Pattern Recognition}.

\bibitem[{Cobbe et~al.(2021)Cobbe, Kosaraju, Bavarian, Chen, Jun, Kaiser, Plappert, Tworek, Hilton, Nakano, Hesse, and Schulman}]{DBLP:journals/corr/abs-2110-14168}
Karl Cobbe, Vineet Kosaraju, Mohammad Bavarian, Mark Chen, Heewoo Jun, Lukasz Kaiser, Matthias Plappert, Jerry Tworek, Jacob Hilton, Reiichiro Nakano, Christopher Hesse, and John Schulman. 2021.
\newblock Training verifiers to solve math word problems.
\newblock \emph{arXiv preprint arXiv:2110.14168}.

\bibitem[{Dao et~al.(2022)Dao, Fu, Ermon, Rudra, and R{\'{e}}}]{DBLP:conf/nips/DaoFERR22}
Tri Dao, Daniel~Y. Fu, Stefano Ermon, Atri Rudra, and Christopher R{\'{e}}. 2022.
\newblock Flashattention: Fast and memory-efficient exact attention with io-awareness.
\newblock In \emph{Advances in Neural Information Processing System}.

\bibitem[{Dettmers et~al.(2022)Dettmers, Lewis, Belkada, and Zettlemoyer}]{dettmers2022llmint88bitmatrixmultiplication}
Tim Dettmers, Mike Lewis, Younes Belkada, and Luke Zettlemoyer. 2022.
\newblock {LLM}.int8(): 8-bit matrix multiplication for transformers at scale.

\bibitem[{Du et~al.(2024)Du, Jiang, Xu, Wu, Yu, Li, Li, Xu, Nie, Tu, and You}]{DBLP:conf/icml/Du0XWY0LXNTY24}
Cunxiao Du, Jing Jiang, Yuanchen Xu, Jiawei Wu, Sicheng Yu, Yongqi Li, Shenggui Li, Kai Xu, Liqiang Nie, Zhaopeng Tu, and Yang You. 2024.
\newblock {G}li{D}e with a cape: {A} low-hassle method to accelerate speculative decoding.
\newblock In \emph{Proceedings of the International Conference on Machine Learning}.

\bibitem[{Du et~al.(2022)Du, Huang, Dai, Tong, Lepikhin, Xu, Krikun et~al.}]{DBLP:conf/icml/DuHDTLXKZYFZFBZ22}
Nan Du, Yanping Huang, Andrew~M. Dai, Simon Tong, Dmitry Lepikhin, Yuanzhong Xu, Maxim Krikun, et~al. 2022.
\newblock {GLaM}: Efficient scaling of language models with mixture-of-experts.
\newblock In \emph{Proceedings of the International Conference on Machine Learning}.

\bibitem[{Fedus et~al.(2022)Fedus, Zoph, and Shazeer}]{DBLP:journals/jmlr/FedusZS22}
William Fedus, Barret Zoph, and Noam Shazeer. 2022.
\newblock Switch transformers: Scaling to trillion parameter models with simple and efficient sparsity.
\newblock \emph{Journal of Machine Learning Research}.

\bibitem[{Frantar et~al.(2023)Frantar, Ashkboos, Hoefler, and Alistarh}]{frantar2023gptqaccurateposttrainingquantization}
Elias Frantar, Saleh Ashkboos, Torsten Hoefler, and Dan Alistarh. 2023.
\newblock {GPTQ}: Accurate post-training quantization for generative pre-trained transformers.

\bibitem[{Grattafiori et~al.(2024)Grattafiori, Dubey, Jauhri, Pandey, Kadian et~al.}]{grattafiori2024llama3herdmodels}
Aaron Grattafiori, Abhimanyu Dubey, Abhinav Jauhri, Abhinav Pandey, Abhishek Kadian, et~al. 2024.
\newblock The llama 3 herd of models.
\newblock \emph{arXiv preprint arXiv:2407.21783}.

\bibitem[{Gu et~al.(2024)Gu, Dong, Wei, and Huang}]{gu2024minillm}
Yuxian Gu, Li~Dong, Furu Wei, and Minlie Huang. 2024.
\newblock Mini{LLM}: Knowledge distillation of large language models.
\newblock In \emph{Proceedings of the International Conference on Learning Representations}.

\bibitem[{Hoffmann et~al.(2022)Hoffmann, Borgeaud, Mensch, Buchatskaya et~al.}]{hoffmann2022trainingcomputeoptimallargelanguage}
Jordan Hoffmann, Sebastian Borgeaud, Arthur Mensch, Elena Buchatskaya, et~al. 2022.
\newblock Training compute-optimal large language models.
\newblock \emph{arXiv preprint arXiv:2203.15556}.

\bibitem[{Hu and Huang(2024)}]{hu2024accelerated}
Zhengmian Hu and Heng Huang. 2024.
\newblock Accelerated speculative sampling based on tree monte carlo.
\newblock In \emph{Proceedings of the International Conference on Machine Learning}.

\bibitem[{Kaplan et~al.(2020)Kaplan, McCandlish, Henighan, Brown, Chess, Child, Gray, Radford, Wu, and Amodei}]{kaplan2020scalinglawsneurallanguage}
Jared Kaplan, Sam McCandlish, Tom Henighan, Tom~B. Brown, Benjamin Chess, Rewon Child, Scott Gray, Alec Radford, Jeffrey Wu, and Dario Amodei. 2020.
\newblock Scaling laws for neural language models.
\newblock \emph{arXiv preprint arXiv:2001.08361}.

\bibitem[{Ko et~al.(2024)Ko, Kim, Chen, and Yun}]{DBLP:conf/icml/KoKCY24}
Jongwoo Ko, Sungnyun Kim, Tianyi Chen, and Se{-}Young Yun. 2024.
\newblock Distillm: Towards streamlined distillation for large language models.
\newblock In \emph{Proceedings of the International Conference on Machine Learning}.

\bibitem[{Kwiatkowski et~al.(2019)Kwiatkowski, Palomaki, Redfield, Collins, Parikh, Alberti, Epstein, Polosukhin, Devlin, Lee, Toutanova, Jones, Kelcey, Chang, Dai, Uszkoreit, Le, and Petrov}]{nq}
Tom Kwiatkowski, Jennimaria Palomaki, Olivia Redfield, Michael Collins, Ankur~P. Parikh, Chris Alberti, Danielle Epstein, Illia Polosukhin, Jacob Devlin, Kenton Lee, Kristina Toutanova, Llion Jones, Matthew Kelcey, Ming{-}Wei Chang, Andrew~M. Dai, Jakob Uszkoreit, Quoc Le, and Slav Petrov. 2019.
\newblock Natural questions: a benchmark for question answering research.
\newblock \emph{Trans. Assoc. Comput. Linguistics}.

\bibitem[{Kwon et~al.(2023)Kwon, Li, Zhuang, Sheng, Zheng, Yu, Gonzalez, Zhang, and Stoica}]{DBLP:conf/sosp/KwonLZ0ZY0ZS23}
Woosuk Kwon, Zhuohan Li, Siyuan Zhuang, Ying Sheng, Lianmin Zheng, Cody~Hao Yu, Joseph Gonzalez, Hao Zhang, and Ion Stoica. 2023.
\newblock Efficient memory management for large language model serving with pagedattention.
\newblock In \emph{Proceedings of the29th Symposium on Operating Systems Principles}.

\bibitem[{Leviathan et~al.(2023)Leviathan, Kalman, and Matias}]{DBLP:conf/icml/LeviathanKM23}
Yaniv Leviathan, Matan Kalman, and Yossi Matias. 2023.
\newblock Fast inference from transformers via speculative decoding.
\newblock In \emph{Proceedings of the International Conference on Machine Learning}.

\bibitem[{Li et~al.(2024{\natexlab{a}})Li, Wei, Zhang, and Zhang}]{li2024eagle2}
Yuhui Li, Fangyun Wei, Chao Zhang, and Hongyang Zhang. 2024{\natexlab{a}}.
\newblock {EAGLE-2}: Faster inference of language models with dynamic draft trees.
\newblock In \emph{Proceedings of the Conference on the Empirical Methods in Natural Language Processing}.

\bibitem[{Li et~al.(2024{\natexlab{b}})Li, Wei, Zhang, and Zhang}]{li2024eagle}
Yuhui Li, Fangyun Wei, Chao Zhang, and Hongyang Zhang. 2024{\natexlab{b}}.
\newblock {EAGLE}: Speculative sampling requires rethinking feature uncertainty.
\newblock In \emph{Proceedings of the International Conference on Machine Learning}.

\bibitem[{Lin et~al.(2024)Lin, Tang, Tang, Yang, Chen, Wang, Xiao, Dang, Gan, and Han}]{DBLP:conf/mlsys/0002TTYCWXDG024}
Ji~Lin, Jiaming Tang, Haotian Tang, Shang Yang, Wei{-}Ming Chen, Wei{-}Chen Wang, Guangxuan Xiao, Xingyu Dang, Chuang Gan, and Song Han. 2024.
\newblock {AWQ:} activation-aware weight quantization for on-device {LLM} compression and acceleration.
\newblock In \emph{Proceedings of the Annual Conference on Machine Learning and Systems}.

\bibitem[{Miao et~al.(2024)Miao, Oliaro, Zhang, Cheng, Wang, engxin Zhang, Wong, Zhu, Yang, Shi, Shi, Chen, Arfeen, Abhyankar, and Jia}]{DBLP:conf/asplos/MiaoOZCWZWZYSSC24}
Xupeng Miao, Gabriele Oliaro, Zhihao Zhang, Xinhao Cheng, Zeyu Wang, engxin Zhang, Rae Ying~Yee Wong, Alan Zhu, Lijie Yang, Xiaoxiang Shi, Chunan Shi, Zhuoming Chen, Daiyaan Arfeen, Reyna Abhyankar, and Zhihao Jia. 2024.
\newblock {S}pec{I}nfer: Accelerating large language model serving with tree-based speculative inference and verification.
\newblock In \emph{Proceedings of the ACM International Conference on Architectural Support for Programming Languages and Operating Systems}.

\bibitem[{Nallapati et~al.(2016)Nallapati, Zhou, dos Santos, G{\"{u}}l{\c{c}}ehre, and Xiang}]{cnndm}
Ramesh Nallapati, Bowen Zhou, C{\'{\i}}cero~Nogueira dos Santos, {\c{C}}aglar G{\"{u}}l{\c{c}}ehre, and Bing Xiang. 2016.
\newblock Abstractive text summarization using sequence-to-sequence rnns and beyond.
\newblock In \emph{Proceedings of the 20th {SIGNLL} Conference on Computational Natural Language Learning, CoNLL 2016, Berlin, Germany, August 11-12, 2016}.

\bibitem[{OpenAI(2023)}]{DBLP:journals/corr/abs-2303-08774}
OpenAI. 2023.
\newblock {GPT-4} technical report.
\newblock \emph{arXiv preprint arXiv:2303.08774}.

\bibitem[{Schroff et~al.(2015)Schroff, Kalenichenko, and Philbin}]{DBLP:conf/cvpr/SchroffKP15}
Florian Schroff, Dmitry Kalenichenko, and James Philbin. 2015.
\newblock {F}ace{N}et: {A} unified embedding for face recognition and clustering.
\newblock In \emph{Proceedings of the Conference on Computer Vision and Pattern Recognition}.

\bibitem[{Sennrich et~al.(2016)Sennrich, Haddow, and Birch}]{sennrich-etal-2016-neural}
Rico Sennrich, Barry Haddow, and Alexandra Birch. 2016.
\newblock Neural machine translation of rare words with subword units.
\newblock In \emph{Proceedings of the Annual Meeting of the Association for Computational Linguistics}.

\bibitem[{Sun et~al.(2024)Sun, Mendlovic, Leviathan, Aharoni, Beirami, Ro, and Suresh}]{sun2024blockverificationacceleratesspeculative}
Ziteng Sun, Uri Mendlovic, Yaniv Leviathan, Asaf Aharoni, Ahmad Beirami, Jae~Hun Ro, and Ananda~Theertha Suresh. 2024.
\newblock Block verification accelerates speculative decoding.
\newblock \emph{arXiv preprint arXiv:2403.10444}.

\bibitem[{Tao et~al.(2024)Tao, Liu, Dou, Muennighoff, Wan, Luo, Lin, and Wong}]{tao2024scalinglawsvocabularylarger}
Chaofan Tao, Qian Liu, Longxu Dou, Niklas Muennighoff, Zhongwei Wan, Ping Luo, Min Lin, and Ngai Wong. 2024.
\newblock Scaling laws with vocabulary: Larger models deserve larger vocabularies.

\bibitem[{Taori et~al.(2023)Taori, Gulrajani, Zhang, Dubois, Li, Guestrin, Liang, and Hashimoto}]{alpaca}
Rohan Taori, Ishaan Gulrajani, Tianyi Zhang, Yann Dubois, Xuechen Li, Carlos Guestrin, Percy Liang, and Tatsunori~B. Hashimoto. 2023.
\newblock Stanford alpaca: An instruction-following llama model.

\bibitem[{Touvron et~al.(2023{\natexlab{a}})Touvron, Lavril, Izacard, Martinet, Lachaux, Lacroix, Rozière, Goyal, Hambro, Azhar, Rodriguez, Joulin, Grave, and Lample}]{touvron2023llamaopenefficientfoundation}
Hugo Touvron, Thibaut Lavril, Gautier Izacard, Xavier Martinet, Marie-Anne Lachaux, Timothée Lacroix, Baptiste Rozière, Naman Goyal, Eric Hambro, Faisal Azhar, Aurelien Rodriguez, Armand Joulin, Edouard Grave, and Guillaume Lample. 2023{\natexlab{a}}.
\newblock Llama: Open and efficient foundation language models.
\newblock \emph{arXiv preprint arXiv:2302.13971}.

\bibitem[{Touvron et~al.(2023{\natexlab{b}})Touvron, Martin, Stone, Albert, Almahairi et~al.}]{touvron2023llama2openfoundation}
Hugo Touvron, Louis Martin, Kevin Stone, Peter Albert, Amjad Almahairi, et~al. 2023{\natexlab{b}}.
\newblock Llama 2: Open foundation and fine-tuned chat models.
\newblock \emph{arXiv preprint arXiv:2307.09288}.

\bibitem[{van~den Oord et~al.(2018)van~den Oord, Li, and Vinyals}]{DBLP:journals/corr/abs-1807-03748}
A{\"{a}}ron van~den Oord, Yazhe Li, and Oriol Vinyals. 2018.
\newblock Representation learning with contrastive predictive coding.
\newblock \emph{arXiv preprint arXiv:1807.03748}.

\bibitem[{Wang et~al.(2020)Wang, Cho, and Gu}]{DBLP:conf/aaai/WangCG20}
Changhan Wang, Kyunghyun Cho, and Jiatao Gu. 2020.
\newblock Neural machine translation with byte-level subwords.
\newblock In \emph{Proceedings of the {AAAI} Conference on Artificial Intelligence}.

\bibitem[{Wang et~al.(2024)Wang, Su, Li, Xia, Ye, Duan, Wang, and Zhang}]{wang2024opttreespeculativedecodingadaptive}
Jikai Wang, Yi~Su, Juntao Li, Qingrong Xia, Zi~Ye, Xinyu Duan, Zhefeng Wang, and Min Zhang. 2024.
\newblock Opt-tree: Speculative decoding with adaptive draft tree structure.
\newblock \emph{arXiv preprint arXiv:2406.17276}.

\bibitem[{Xiao et~al.(2024)Xiao, Shi, Nie, Yang, Deng, Su, Chen, and Cui}]{DBLP:journals/corr/abs-2405-00263}
Bin Xiao, Chunan Shi, Xiaonan Nie, Fan Yang, Xiangwei Deng, Lei Su, Weipeng Chen, and Bin Cui. 2024.
\newblock Clover: Regressive lightweight speculative decoding with sequential knowledge.
\newblock \emph{arXiv preprint arXiv:2405.00263}.

\bibitem[{Xiao et~al.(2023)Xiao, Lin, Seznec, Wu, Demouth, and Han}]{DBLP:conf/icml/XiaoLSWDH23}
Guangxuan Xiao, Ji~Lin, Micka{\"{e}}l Seznec, Hao Wu, Julien Demouth, and Song Han. 2023.
\newblock {S}mooth{Q}uant: Accurate and efficient post-training quantization for large language models.
\newblock In \emph{Proceedings of the International Conference on Machine Learning}.

\bibitem[{Yang et~al.(2024)Yang, Yang, Zhang, Hui et~al.}]{qwen2.5}
An~Yang, Baosong Yang, Beichen Zhang, Binyuan Hui, et~al. 2024.
\newblock Qwen2.5 technical report.
\newblock \emph{arXiv preprint arXiv:2412.15115}.

\bibitem[{Zhang et~al.(2024)Zhang, Wang, Huang, and Xu}]{zhang2024learningharmonizedrepresentationsspeculative}
Lefan Zhang, Xiaodan Wang, Yanhua Huang, and Ruiwen Xu. 2024.
\newblock Learning harmonized representations for speculative sampling.
\newblock \emph{arXiv preprint arXiv:2408.15766}.

\bibitem[{Zheng et~al.(2023)Zheng, Chiang, Sheng, Zhuang, Wu, Zhuang, Lin, Li, Li, Xing, Zhang, Gonzalez, and Stoica}]{DBLP:conf/nips/ZhengC00WZL0LXZ23}
Lianmin Zheng, Wei{-}Lin Chiang, Ying Sheng, Siyuan Zhuang, Zhanghao Wu, Yonghao Zhuang, Zi~Lin, Zhuohan Li, Dacheng Li, Eric~P. Xing, Hao Zhang, Joseph~E. Gonzalez, and Ion Stoica. 2023.
\newblock Judging llm-as-a-judge with mt-bench and chatbot arena.
\newblock In \emph{Advances in Neural Information Processing Systems}.

\bibitem[{Zhong et~al.(2024)Zhong, Ding, Shen, Liu, Du, and Tao}]{DBLP:conf/acl/Zhong00L0T24}
Qihuang Zhong, Liang Ding, Li~Shen, Juhua Liu, Bo~Du, and Dacheng Tao. 2024.
\newblock Revisiting knowledge distillation for autoregressive language models.
\newblock In \emph{Proceedings of the Annual Meeting of the Association for Computational Linguistics}.

\end{thebibliography}

\appendix
\section{Hyperparameters in CSRA Loss}
\label{sec:appendixA}
The temperature of $\mathcal{L}_{CSRA}$ is set to 0.07, consistent with some previous works such as CLIP \cite{DBLP:conf/nips/ZhengC00WZL0LXZ23}.

Then we set $w_{reg}$ to 0.5, half of EAGLE's original setting. The weight of CSRA loss is adjusted according to different target models, making the values of $w_{CSRA}\mathcal{L}_{CSRA}$ and $w_{reg}\mathcal{L}_{reg}$ roughly the same. In this way, the loss imposed on representation is approximately the same as EAGLE/HASS training.

Based on the values of $w_{reg}\mathcal{L}_{reg}$, we choose $w_{CSRA}=0.2$ for Qwen2.5-7B, $w_{CSRA}=0.15$ for Llama3-8B, $w_{CSRA}=0.1$ for Llama3-70B, Qwen2.5-14B and Llama2-7B, and 0.05 for Llama2-13B.

\section{Training Details}
\label{sec:appendixE}
We use a fixed dataset of 68,000 examples from ShareGPT\footnote{https://huggingface.co/datasets/Aeala/ShareGPT\_Vicuna\\\_unfiltered} as our training set, which is identical to EAGLE and HASS. CORAL requires approximately 2 days to train a 7B draft model under default settings (training step=3, epoch=20). It is worth noting that draft models with large vocabularies such as Llama3 and Qwen2.5 require more GPU memory compared to Llama2, so we use 4$\times$NVIDIA H20-96G GPUs for training. Training large draft models such as Llama3-70B on A100-40G GPU may result in out-of-memory issues under our experimental settings. We recommend using GPUs with larger memory capacities or choosing other alternatives (\emph{e.g.}, reducing the batch size, model parallelism).

\section{Single-step Training with CSRA}
\label{sec:appendixD}
We do not recommend using the CSRA loss in the context of single-step training. Our empirical findings suggest that introducing the CSRA loss may lead to a slight increase in regression loss, likely due to the mismatch between the two optimization objectives. Specifically, the CSRA loss focuses solely on the angular relationships between the output features, without imposing any constraints on the feature norm, whereas the regression loss aims to learn features that are identical to the target. The increase in regression loss may damage the acceptance length. We present the results of CSRA with single-step training in Table \ref{tab:single_step}.

\begin{table}[h]
    \centering
    \setlength{\tabcolsep}{0.12cm}
    \begin{tabular}{cccc}
    \hline
        ~ & MT-bench & HumanEval & GSM8K \\ \hline
        EAGLE-2 & 4.16 & 4.78 & 4.21 \\
        CSRA Step1 & 4.10 & 4.70 & 4.10 \\ \hline
    \end{tabular}
    \caption{Acceptance length of Llama3-8B EAGLE-2 and CORAL model with single-step training.}
    \label{tab:single_step}
\end{table}

A plausible explanation for this phenomenon is that in single-step training, the draft model lacks exposure to subsequent steps, therefore the L1 distance between the prediction and target feature is relatively more critical. In contrast, for multi-step training, the draft model learns to adapt to subsequent steps, making the discriminative power of different representations and the multi-step consistency more crucial.

\section{Discussion on the discrepancies between different training steps}
\label{sec:appendix_CSRA}
To better illustrate the discrepancies between representations from multiple training steps, we measure the InfoNCE between features from different steps. Please note that absolute distance metrics (such as L1 or cosine distance) are not ideal measurements, as absolute distances fail to represent the distinguishability between different features. In contrast, InfoNCE transforms cosine similarity into a probability distribution, effectively reflecting the relative distances between features, which is more crucial for prediction accuracy. Therefore, InfoNCE serves as a more appropriate metric.

\begin{table}[ht]
    \centering
    \setlength{\tabcolsep}{0.12cm}
    \begin{tabular}{c|cccc}
    \toprule       
        EAGLE-2 & step 0 & step 1 & step 2 & step 3 \\ \hline
        step 0 & - & - & - & - \\ 
        step 1 & 1.5668 & - & - & - \\ 
        step 2 & 1.8415 & 1.4843 & - & - \\ 
        step 3 & 2.0391 & 1.6559 & 1.4876 & - \\ \midrule
        HASS & step 0 & step 1 & step 2 & step 3 \\ \hline
        step 0 & - & - & - & - \\ 
        step 1 & 1.4577 & - & - & - \\ 
        step 2 & 1.6117 & 1.3816 & - & - \\ 
        step 3 & 1.7290 & 1.4876 & 1.3924 & - \\ \midrule
        CSRA & step 0 & step 1 & step 2 & step 3 \\ \hline
        step 0 & - & - & - & - \\ 
        step 1 & 0.9545 & - & - & - \\ 
        step 2 & 1.1044 & 0.8395 & - & - \\ 
        step 3 & 1.2179 & 0.9381 & 0.8173 & - \\ \bottomrule
    \end{tabular}
    \caption{InfoNCE between features from different training steps. The temperature is set to 0.07, which is aligned with our setting in CSRA training.}
\end{table}

We randomly select 100 samples from ShareGPT and evaluate the differences in output features across 4 steps. The following are the InfoNCE between features of different steps for EAGLE-2, HASS, and CSRA. For instance, the first column represents the InfoNCE between the output features of step 1 and step 2 to 4. Clearly, the differences in features between steps increase gradually as the number of steps grows. Since HASS employs multi-step training, the differences between steps are smaller compared to EAGLE-2. Moreover, our method significantly reduces the discrepancies between different steps, achieves higher similarity between positive features and enhances the discriminative power of negative features, ensuring relatively consistent performance across all steps during inference.

\begin{table*}[t]
    \centering
    \begin{tabular}{cc|ccc}
    \hline
        Train & Test & MT-bench & HumanEval & GSM8K \\ \hline
        \multirow{2}*{\makecell{sys\_p2}} & sys\_p2 & 4.16 & 4.78 & 4.21 \\ 
        ~ & sys\_p1 & 4.11(\textcolor{red}{-0.05}) & 4.73(\textcolor{red}{-0.05}) & 4.27(\textcolor{green}{+0.06}) \\ \hline
        \multirow{2}*{\makecell{sys\_p1}} & sys\_p1 & 4.18 & 4.78 & 4.38 \\ 
        ~ & sys\_p2 & 3.87(\textcolor{red}{-0.31}) & 4.17(\textcolor{red}{-0.61}) & 3.93(\textcolor{red}{-0.45}) \\ \hline
        \multirow{2}*{\makecell{open source \\ (sys\_p1)}} & sys\_p1 & 4.24 & 4.92 & 4.34 \\ 
        ~ & sys\_p2 & 3.94(\textcolor{red}{-0.30}) & 4.67(\textcolor{red}{-0.25}) & 3.91(\textcolor{red}{-0.43}) \\ \hline
    \end{tabular}
    \caption{Acceptance lengths of EAGLE-2 for Llama3-8B-Instruct with different system prompts.}
    \label{tab:sys_p}
\end{table*}

\section{Discussion on System Prompt}
\label{sec:appendixC}
EAGLE utilizes the system prompt from the official Llama2-chat example\footnote{https://huggingface.co/blog/llama2}:

\textcolor{blue}{sys\_p1 = You are a helpful, respectful and honest assistant. Always answer as helpfully as possible, while being safe.  Your answers should not include any harmful, unethical, racist, sexist, toxic, dangerous, or illegal content. Please ensure that your responses are socially unbiased and positive in nature.\textbackslash n\textbackslash nIf a question does not make any sense, or is not factually coherent, explain why instead of answering something not correct. If you don't know the answer to a question, please don't share false information.}

The same system prompt is also used in Llama3 drafter training. However, it appears that Llama3 does not have a default system prompt. Nevertheless, we find the system prompt in the official Llama3.3 example\footnote{https://github.com/meta-llama/llama-models/blob/main/models/llama3\_3/prompt\_format.md} is simpler and also widely adopted:

\textcolor{red}{sys\_p2 = You are a helpful assistant}

The system prompt has a certain impact on the acceptance length and speedup ratio. To investigate this, we compared the open-source Llama3-8B-Instrct draft model in EAGLE official repository (trained with \textcolor{blue}{sys\_p1}) and draft models trained by ourselves using \textcolor{blue}{sys\_p1} and \textcolor{red}{sys\_p2}. Our results in Table \ref{tab:sys_p} show that switching between different system prompts might lead to a decrease in speedup and acceptance length on the MT-Bench and Humaneval datasets, while GSM8K is an exception.

Upon closer inspection of the GSM8K results, we find that when using sys\_p1, most responses start with a sentence similar to "Let's break this down step by step", whereas when using sys\_p2, the beginning if outputs will be more diverse. This suggests that the speedup ratio using sys\_p1 might be artificially inflated in some cases.

Furthermore, since longer system prompts provide the draft model with more context, we suppose that detailed prompts and increased information could potentially improve the performance of draft model when the system prompt of training and inference is aligned. However, when the system prompts are not consistent, training the model with a more detailed system prompt may lead to greater performance degradation.

To obtain a more generalizable draft model, we use \textcolor{red}{sys\_p2} in all experiments with Llama3-Instruct 8B/70B. We believe a more general and simple system prompt would reflect the draft model's true capabilities more accurately.

\section{Discussion on LM Head Router}
\label{sec:appendixB}
In this section, we will discuss some issues of LM head router. 

\noindent\textbf{Tree decoding.} In tree decoding, each timestep contains multiple candidate tokens. Since each candidate requires a different set of LM head groups, we need to activate all the involved groups, which may bring additional latency. In some cases, we even need to activate the entire LM head parameters (\emph{e.g.}, if we take the top two groups and top 10 candidates, the worst-case scenario might require activating 20 groups). 

This issue can be addressed through appropriate grouping strategies. First, dividing the tokens into more groups helps alleviate the problem. For instance, with a total of 32 groups, selecting the top 10 candidates from the top 2 groups ensures that the LM head parameters are not fully activated, even in the worst-case scenario. Second, modern LLMs utilize BPE \cite{sennrich-etal-2016-neural} or BBPE \cite{DBLP:conf/aaai/WangCG20} for tokenization, where higher-frequency tokens tend to be concentrated in groups with smaller indices. As a result, such an extreme scenario is unlikely to occur in practice.

\noindent\textbf{Two-stage training.} There are mainly two reasons for adopting two-stage training. Firstly, the two-stage training strategy ensures that the router serves as a plug-and-play module, without affecting the standalone usage of the first-stage model, thereby providing greater flexibility. Secondly, since the number of groups is a hyperparameter that may require multiple experiments to determine the optimal setting, two-stage training allows us to store the output of draft model and train the router only, making it easier for parameter tuning. 

\begin{table*}[t]
    \centering
    \begin{tabular}{cc|cc|cc|cc} \hline
        ~ & ~ & \multicolumn{2}{c|}{Alpaca} & \multicolumn{2}{c|}{Natural Q.} & \multicolumn{2}{c}{CNN/DM} \\ 
        model & method & $\tau$ & SR & $\tau$ & SR & $\tau$ & SR \\ \hline
        \multirow{3}*{\makecell{L2-7B}} & EAGLE-2 & 4.51 & 2.88$\times$ & 4.10 & 2.61$\times$ & 4.12 & 2.40$\times$ \\
        ~ & HASS & 4.87 & 3.11$\times$ & 4.41 & 2.80$\times$ & 4.44 & 2.57$\times$ \\ 
        \rowcolor{gray!10} \cellcolor{white} ~ & CORAL & \textbf{4.96} & \textbf{3.15$\times$} & \textbf{4.44} & \textbf{2.84$\times$} & \textbf{4.54} & \textbf{2.62$\times$} \\ \hline
        \multirow{4}*{\makecell{L3-8B}} & EAGLE-2 & 4.33 & 2.39$\times$ & 3.37 & 1.86$\times$ & 3.82 & 1.98$\times$ \\
        ~ & HASS & 4.77 & 2.56$\times$ & 3.59 & 1.98$\times$ & 4.06 & 2.16$\times$ \\ 
        ~ & \cellcolor{gray!10}CORAL & \cellcolor{gray!10}\textbf{4.79} & \cellcolor{gray!10}\textbf{2.58$\times$} & \cellcolor{gray!10}\textbf{3.63} & \cellcolor{gray!10}\textbf{2.00$\times$} & \cellcolor{gray!10}\textbf{4.16} & \cellcolor{gray!10}\textbf{2.20$\times$} \\ 
        ~ & \cellcolor{gray!20}CORAL w/ r. & \cellcolor{gray!20}4.49 & \cellcolor{gray!20}\underline{\textbf{2.74$\times$}} & \cellcolor{gray!20}3.28 & \cellcolor{gray!20}\underline{\textbf{2.06$\times$}} & \cellcolor{gray!20}3.61 & \cellcolor{gray!20}2.16$\times$ \\ \hline
        \multirow{4}*{\makecell{Q2.5-7B}} & EAGLE-2 & 3.93 & 2.17$\times$ & 3.13 & 1.73$\times$ & 3.33 & 1.78$\times$ \\
        ~ & HASS & 4.19 & 2.31$\times$ & 3.30 & 1.82$\times$ & 3.57 & 1.90$\times$ \\
        ~ & \cellcolor{gray!10} CORAL & \cellcolor{gray!10}\textbf{4.29} & \cellcolor{gray!10}\textbf{2.35$\times$} & \cellcolor{gray!10}\textbf{3.38} & \cellcolor{gray!10}\textbf{1.86$\times$} & \cellcolor{gray!10}\textbf{3.72} & \cellcolor{gray!10}\textbf{1.97$\times$} \\
        ~ & \cellcolor{gray!20}CORAL w/ r. & \cellcolor{gray!20}4.11 & \cellcolor{gray!20}\underline{\textbf{2.60$\times$}} & \cellcolor{gray!20}3.15 & \cellcolor{gray!20}\underline{\textbf{1.99$\times$}} & \cellcolor{gray!20}3.28 & \cellcolor{gray!20}\underline{\textbf{1.99$\times$}} \\ \hline
    \end{tabular}
    \caption{Additional results on Alpaca, Natural Questions and CNN/DM dataset. We provide the results on Llama2-7B-chat, Llama3-8B-Instruct and Qwen2.5-7B-Instruct. The temperature is set to 0.}
    \label{tab:additional results}
\end{table*}

\noindent\textbf{Backends.}
Although many researches on speculative decoding measure the speedup ratio on PyTorch, we do not consider PyTorch to be a good backend. For example, as shown in Table \ref{fig:pie}, the FP16 latency of Llama3-8B-draft head on RTX A6000 GPU is 1.51ms, which is close to the theoretical time of 1.3ms (1002M memory access with 768GB/s bandwidth). However, for other parts, which mainly consists of transformer, the actual time is much higher than the theoretical time (1.07ms vs 0.63ms), achieving only about 60\% of the theoretical performance.

This is a problem inherent to PyTorch. For instance, in Qwen2 speed benchmark\footnote{https://qwen.readthedocs.io/en/v2.0/benchmark/speed\_be\\nchmark.html}, the inference speed of 7B model on A100 80G GPU is only 38 token/s (\emph{i.e.}, 26ms/token), which is far from the theoretical time of about 7ms (estimated by 14G memory access with 2TB/s bandwidth). This problem can be mitigated by using a more optimized backend, such as vLLM \cite{DBLP:conf/sosp/KwonLZ0ZY0ZS23}.

Therefore, the performance of the LM head router may be affected by the hardware and backend conditions. In a well-optimized backend, the router's performance will be better than reported in this paper, as the latency of the LM head will occupy a larger proportion in the draft model.

\noindent\textbf{Small vocabulary and super large LLMs.} Let’s take Llama3-70B on MT-bench as an example. In our experiments on CORAL, the time consumption of the target/draft model is 4105s/561s, meaning that the draft model accounts for only 12\% of the entire drafting-verification cycle (for Llama3-8B, this figure is approximately 33\%). Although the LM head of the draft model still constitutes a significant portion of the drafting latency, its overall contribution to the entire cycle is only 5–6\% (while for Llama3-8B, it is nearly 20\%). If a router is used, the time consumption of the target/draft model becomes 4477s/440s, resulting in only a 3\% reduction in the entire cycle. However, the acceptance length decreases from 4.23 to 3.93, a drop of 9.3\%, and the speedup decreases from 2.79$\times$ to 2.69$\times$.

A similar conclusion applies to Llama2-7B. Since the latency of the LM head does not constitute a large part of the total latency, using a router on Llama2 is not a good choice.

\section{Additional Experiments}
Here we present some additional experimental results on Alpaca \cite{alpaca}, Natural Questions \cite{nq} and CNN/DM \cite{cnndm} datasets.

\section{Licenses of Artifacts}
\label{sec:appendixF}
We present the licenses of artifacts related to this paper in table \ref{tab:license}.
\begin{table}[htb]
    \centering
    \begin{tabular}{c|c|c}
    \toprule
        \multirow{3}*{\makecell{models}} & Llama3 & llama3 license \\ 
        ~ & Llama2 & llama2 license \\ 
        ~ & Qwen2.5 & apache-2.0 \\ \midrule
        \multirow{7}*{\makecell{datasets}} & ShareGPT & apache-2.0 \\ 
        ~ & MT-bench & CC-BY-4.0 \\ 
        ~ & HumanEval & MIT \\ 
        ~ & GSM8K & MIT \\ 
        ~ & Alpaca & CC-BY-NC-4.0 \\ 
        ~ & Natural Questions & apache-2.0 \\
        ~ & CNN/DM & apache-2.0 \\ \midrule
        \multirow{2}*{\makecell{codes}} & EAGLE/EAGLE2 & apache-2.0 \\ 
        ~ & HASS & not provided \\ \bottomrule
    \end{tabular}
    \caption{Licenses of artifacts}
    \label{tab:license}
\end{table}

\end{document}